\documentclass{article}

 \usepackage[dblblindworkshop, final]{neurips_2025}
\workshoptitle{Space in Vision, Language, and Embodied AI}

\usepackage[utf8]{inputenc} 
\usepackage[T1]{fontenc}    
\usepackage{hyperref}       
\usepackage{url}            
\usepackage{booktabs}       
\usepackage{amsfonts}       
\usepackage{nicefrac}       
\usepackage{microtype}      
\usepackage{xcolor}         

\usepackage{url}
\usepackage[utf8]{inputenc} 
\usepackage[T1]{fontenc}    
\usepackage{booktabs}       
\usepackage{amsfonts}       
\usepackage{nicefrac}       
\usepackage{microtype}      
\usepackage{xcolor}         
\usepackage{adjustbox}
\usepackage{multirow}
\usepackage{caption}
\usepackage{subcaption}
\usepackage{tablefootnote}
\usepackage{amsmath}
\usepackage{placeins}
\usepackage{cleveref}

\title{NinA: Normalizing Flows in Action. Training VLA Models with Normalizing Flows}

\author{
Denis Tarasov\thanks{Correspondence to: \href{mailto:tarasovd@ethz.ch}{tarasovd@ethz.ch}. Work done by \href{https://dunnolab.ai}{dunnolab.ai}.} \\ AIRI, ETH Zürich \And 
Alexander Nikulin \\ AIRI, MIPT \And 
Ilya Zisman \\ AIRI, Skoltech \And 
Albina Klepach \\ AIRI \And 
Nikita Lyubaykin \\ AIRI, Innopolis University \And 
Andrei Polubarov \\ AIRI, Skoltech \And 
Alexander Derevyagin \\ AIRI, HSE \And 
Vladislav Kurenkov \\ AIRI, Innopolis University
}

\begin{document}

\maketitle

\setcounter{footnote}{0}
\begin{abstract}
  Recent advances in Vision-Language-Action (VLA) models have established a two-component architecture, where a pre-trained Vision-Language Model (VLM) encodes visual observations and task descriptions, and an action decoder maps these representations to continuous actions. Diffusion models have been widely adopted as action decoders due to their ability to model complex, multimodal action distributions. However, they require multiple iterative denoising steps at inference time or downstream techniques to speed up sampling, limiting their practicality in real-world settings where high-frequency control is crucial. In this work, we present NinA (Normalizing Flows in Action), a fast and expressive alternative to diffusion-based decoders for VLAs. NinA replaces the diffusion action decoder with a Normalizing Flow (NF) that enables one-shot sampling through an invertible transformation, significantly reducing inference time. We integrate NinA into the FLOWER VLA architecture and fine-tune on the LIBERO benchmark. Our experiments show that NinA matches the performance of its diffusion-based counterpart under the same training regime, while achieving substantially faster inference. These results suggest that NinA offers a promising path toward efficient, high-frequency VLA control without compromising performance. \footnote{Our code is available at \url{https://github.com/dunnolab/NinA/}}
\end{abstract}

\section{Introduction}

The field of general-purpose robotics has recently seen rapid progress driven by Vision-Language-Action (VLA) models \citep{brohan2022rt, zitkovich2023rt, wu2023unleashing, team2024octo, kim2024openvla, black2410pi0, reussflower}.
These models combine pre-trained Vision-Language Models (VLMs) -- leveraging large-scale multimodal pretraining with an action prediction module that outputs low-level robot commands given a textual goal description and visual observations of the environment.
One particularly effective architectural pattern, first successfully demonstrated in $\pi_0$ \citep{black2410pi0} and subsequently adopted in more advanced systems such as $\pi_{0.5}$ \citep{intelligence2025pi_} and FLOWER \citep{reussflower}, involves splitting the VLA into two components: a frozen or fine-tuned VLM encoder and an action expert that maps VLM embeddings into continuous actions. 
\begin{figure*}
\centering
    \begin{subfigure}[b]{0.99\textwidth}
        \centering
        \centerline{\includegraphics[width=\columnwidth]{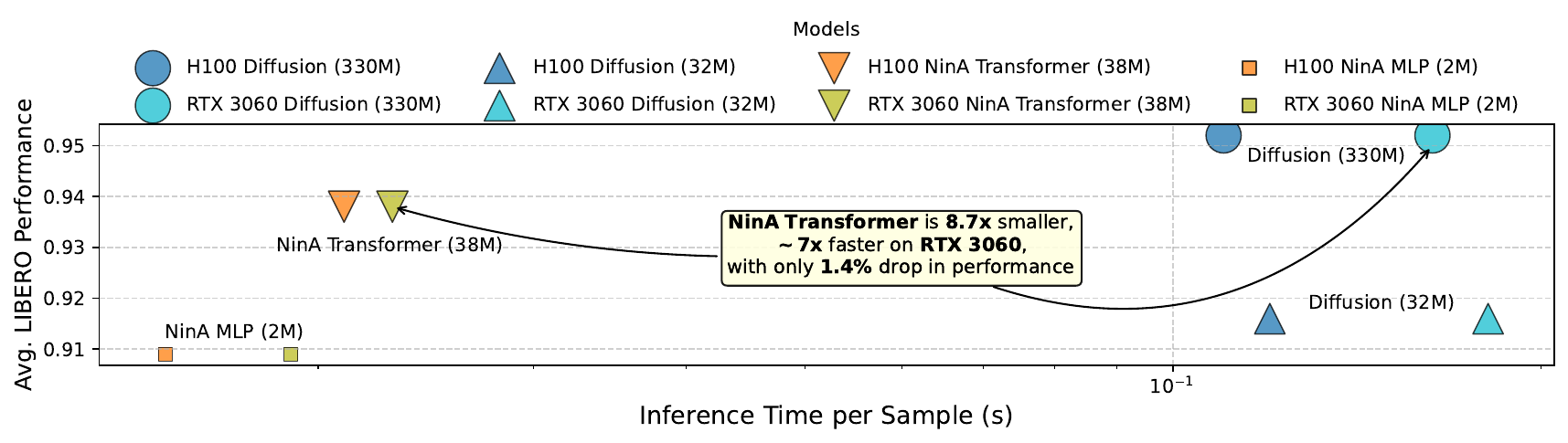}}
        \label{fig:}
    \end{subfigure}
    \caption{Comparison of model performance, size, and inference time on the LIBERO benchmark. Our NinA models (Transformer and MLP) achieve inference speeds up to 10x faster and require significantly fewer parameters compared to diffusion models, while maintaining comparable performance.}
    \label{fig:overall_avg}
\end{figure*}

In prior work, the action expert is almost always implemented using diffusion-based generative models. These models are well-regarded for their ability to capture complex, multi-modal action distributions and have achieved dominant status in many other generative domains such as image and audio synthesis \citep{yang2023diffusion}.
However, their autoregressive denoising process requires multiple forward passes during inference (or specialized acceleration methods that still incur additional computation), leading to latency bottlenecks, a critical concern for fine-grained, real-time robotic control \citep{zhao2023learning}.

We argue that Normalizing Flows (NFs) \citep{dinh2014nice, rezende2015variational} offer a compelling alternative for the action expert. Like diffusion models, NFs can represent complex probability distributions, but they require only a single forward pass to generate an action.
In addition, NFs natively provide exact likelihood estimation and support variational inference, features that could be valuable for downstream tasks such as reinforcement learning (RL), uncertainty estimation, or interpretability \citep{zhai2024normalizing}.
Recent works have also demonstrated the potential of NFs in imitation learning and RL \citep{akimov2022let, ghugare2025normalizing}.

In this preliminary work, we introduce \textbf{N}ormalizing Flows \textbf{in} \textbf{A}ction (\textbf{NinA}) -- a VLA variant that replaces the diffusion-based action expert with a normalizing flow model.
Using the FLOWER VLA \citep{reussflower} as our base architecture, we fine-tune NinA and the original diffusion-based FLOWER on the LIBERO benchmark, finding that NFs can match the performance of state-of-the-art diffusion experts while being significantly faster at inference and requiring fewer parameters.
We also investigate key design choices for NinA, including backbone architecture (MLP vs. Transformer \citep{vaswani2017attention}) and the impact of various hyperparameters, to better understand the trade-offs involved in adopting flows for robotic control.

\section{Preliminaries}
\subsection{Normalizing Flows}
Normalizing Flows (NFs) \citep{dinh2014nice, rezende2015variational} are a family of generative models that represent a complex probability distribution by applying a sequence of invertible transformations to a simple base distribution (e.g., a Gaussian). 
Let $\mathbf{z}_0 \sim p_0(\mathbf{z}_0)$ denote a sample from the base distribution and $f_{\theta} = f_K \circ \dots \circ f_1$ be a composition of $K$ invertible mappings with tractable Jacobian determinants. 
The transformed variable $\mathbf{z}_K = f_{\theta}(\mathbf{z}_0)$ follows the distribution
\begin{equation}
    \label{formula:loss}
    \log p_{\theta}(\mathbf{z}_K) = \log p_0(\mathbf{z}_0) - \sum_{k=1}^{K} \log \left| \det \frac{\partial f_k}{\partial \mathbf{z}_{k-1}} \right|.
\end{equation}
The exact likelihood calculation enables direct maximum likelihood training and facilitates uncertainty estimation, variational inference, and probabilistic reasoning. 
Inference is efficient, requiring only a single forward pass through the sequence of transformations to produce a sample.

\subsection{Vision-Language-Action Models Objective}
Vision-Language-Action (VLA) models extend vision-language models by adding an \emph{action expert} that maps multimodal representations to continuous control commands. 
Given a visual observation $\mathbf{o}_t$ and a textual instruction $\mathbf{g}$, a VLA model first encodes these inputs into a joint embedding $\mathbf{h}_t$ using a pre-trained or fine-tuned vision-language backbone:
\begin{equation}
    \mathbf{h}_t = \mathrm{VLM}(\mathbf{o}_t, \mathbf{g}).
\end{equation}
The action expert $\pi_{\theta}$ then generates an action or a chunk of actions $\mathbf{a}_t \sim \pi_{\theta}(\cdot \mid \mathbf{h}_t)$ in the robot action space. 
For imitation learning, the objective is to maximize the log-likelihood of expert actions:
\begin{equation}
    \mathcal{L}_{\mathrm{VLA}}(\theta) = \mathbb{E}_{(\mathbf{o}_t, \mathbf{g}, \mathbf{a}_t) \sim \mathcal{D}} 
    \left[ \log \pi_{\theta}(\mathbf{a}_t \mid \mathrm{VLM}(\mathbf{o}_t, \mathbf{g})) \right],
\end{equation}
where $\mathcal{D}$ is a dataset of demonstration trajectories.
This modular design enables one to swap different action experts, such as diffusion models or normalizing flows, while reusing the same vision-language backbone.

\section{Methodology}
\label{methodology}
NinA implements Normalizing Flows (NFs) similar to RealNVP \citep{dinh2016density} in two variants: an MLP-based architecture inspired by \citet{ghugare2025normalizing} and a Transformer-based architecture inspired by Jet~\citep{kolesnikov2024jet}. The MLP variant enables faster inference and reduced memory consumption, while the Transformer variant offers better performance and scalability.

During training, state–goal–action chunks $(\mathbf{o}_t, \mathbf{g}, \mathbf{a}_t)$ are sampled from the dataset $\mathcal{D}$. The state $\mathbf{o}_t$ and goal $\mathbf{g}$ are passed through a pretrained VLM to obtain embeddings $\mathbf{h}_t$.

We add Gaussian noise $\mathcal{N}(0, \sigma^2_{\mathrm{noise}})$ to the action chunks, following the beneficial practice for NFs proposed by \citet{zhai2024normalizing}. While this technique has not been nicely ablated in the context of continuous control \citep{ghugare2025normalizing}, our experiments confirm its utility. We also find that the noise amplitude is a critical hyperparameter and report an ablation study on its effect.

The noised actions $\mathbf{\hat{a}}_t$ (treated as $\mathbf{z}_K$) are passed through a sequence of flow layers $f_k$. In each layer, $\mathbf{a}_t$ is randomly split into two equal parts, $x_1$ and $x_2$. For the MLP-based approach, $\mathbf{\hat{a}}_t$ is treated as a single real-valued vector and split element-wise. For the Transformer-based model, $\mathbf{\hat{a}}_t$ is treated as a sequence, and the split is performed sequence-wise, meaning that each action appears fully in either $x_1$ or $x_2$.

Next, $x_1$ is passed through a trainable network $g_{\phi_k}(x_1, \mathbf{h}_t)$ to produce scale and bias terms $(s, b)$, conditioned on the VLM output. In the MLP variant, $g_{\phi_k}$ is an MLP, with conditioning achieved by concatenating $x_1$ and $\mathbf{h}_t$ into a single input vector. In the Transformer variant, $g_{\phi_k}$ consists of stacked self- and cross-attention layers, with conditioning performed via the cross-attention mechanism.

Following \citet{kolesnikov2024jet}, we apply a $\tanh$ activation to $s$ to stabilize training. The second part, $x_2$, is then transformed as:
\begin{equation}
y_2 = \exp(s) \cdot x_2 + b.
\end{equation}
Finally, $y_2$ is concatenated with $x_1$ to produce the output $\mathbf{z}'_k$ of the coupling layer.

Following \citet{ghugare2025normalizing}, we also include the trainable invertible linear layer $PLU_k$ from \citet{kingma2018glow}, applied to $\mathbf{z}'_k$ to produce the final output $\mathbf{z}_k$ of the $k$-th flow layer. While the integration was straightforward for the MLP variant, applying $PLU_k$ action-wise in the Transformer variant caused divergence. This was resolved by applying $PLU_k$ to the entire chunk, treating it as a single vector.

The final latent variable $\mathbf{z}_0$ is used to compute the log-probability under $p_0$, which in our experiments is set to the standard normal distribution $\mathcal{N}(0, I)$, a common choice in NF literature. The design of the transformations allows efficient computation of the Jacobian determinant $\det \frac{\partial f_k}{\partial \mathbf{z}_{k-1}}$, enabling direct computation of the loss in \autoref{formula:loss}.

At inference time, $\mathbf{z}_0$ is sampled from $p_0$ and passed in reverse through the flow layers using the inverse transformations, producing a chunk of actions. All other components remain identical to FLOWER~\citep{reussflower}, with NinA replacing the original diffusion-based action expert. A visual illustration of the architecture is shown in \autoref{app:additional}.

\section{Experimental Results}

\begin{table}
    \begin{center}
    \caption{Performance of NinA compared to diffusion-based experts on the LIBERO benchmark. We report success rates across all LIBERO variants and their average. While the original large diffusion policy achieves the highest average score, NinA models achieve comparable performance with drastically fewer parameters (2M–38M vs. 31M–330M) and much faster inference.}

    \begin{small}
    \begin{adjustbox}{max width=\columnwidth}
    \label{tab:core}
				\begin{tabular}{l|ccccc|c}
		\toprule
	\textbf{Model} & \textbf{LIBERO Spatial} & \textbf{LIBERO Object} & \textbf{LIBERO Goal} & \textbf{LIBERO 10} & \textbf{LIBERO 90} & \textbf{Avg.}\\
\midrule
\textbf{Diffusion (330M, Original)} & 0.982 & 0.976 & 0.942 & 0.906 & 0.954 & 0.952 \\
\textit{-- No robotic VLM pretrain} & 0.986 & 0.924 & 0.980 & 0.896 & 0.941 & 0.945 \\
\midrule
\textbf{Diffusion (31M)} & 0.890 & 0.984 & 0.952 & 0.864 & 0.894 & 0.916 \\
\midrule
\textbf{NinA MLP (2M)} & 0.878 & 0.982 & 0.902 & 0.928 & 0.856 & 0.909 \\
\textit{-- No robotic VLM pretrain} & 0.940 & 0.982 & 0.938 & 0.894 & 0.857 & 0.922 \\
\textit{-- No PLU} & 0.872 & 0.992 & 0.960 & 0.880 & 0.852 & 0.911 \\
\textit{-- No noise} & 0.846 & 0.968 & 0.902 & 0.898 & 0.790 & 0.880 \\
\midrule
\textbf{NinA Transformer (38M)} & 0.970 & 0.978 & 0.938 & 0.920 & 0.887 & 0.938 \\
\textit{-- No robotic VLM pretrain} & 0.960 & 0.976 & 0.926 & 0.908 & 0.895 & 0.933 \\
\textit{-- No PLU} & 0.948 & 0.972 & 0.944 & 0.920 & 0.887 & 0.934 \\
\textit{-- No noise} & 0.960 & 0.980 & 0.888 & 0.850 & 0.803 & 0.896 \\
\midrule
\end{tabular}
        \end{adjustbox}
    \end{small}
    \end{center}
    \vskip -0.1in
\end{table}

We evaluate NinA on the LIBERO benchmark \citep{liu2023libero}, following the fine-tuning protocol of FLOWER \citep{reussflower}. As our primary baseline, we use FLOWER with its original diffusion-based policy. For fairness, we reinitialize the action expert in all experiments, as reproducing the full pretraining of FLOWER is computationally prohibitive. To identify the best hyperparameters for NinA, we tuned exclusively on the LIBERO 10 task suite and applied the selected configuration to all other LIBERO tasks. While this choice limits task-specific optimization, it provides a fairer comparison and avoids overfitting to individual benchmarks. Additional training details are provided in \autoref{app:details}. To disentangle the effect of model capacity, we further downscaled the diffusion baseline to approximately match the size of our best NinA Transformer (38M parameters), resulting in a 31M-parameter diffusion expert. We did not attempt to reduce diffusion models to the scale of NinA MLP (2M parameters), as even the 31M variant already suffered a substantial performance drop.

\autoref{tab:core} summarizes the results across all LIBERO variants. The original diffusion policy achieves the highest average success rate, but NinA delivers competitive performance while being significantly more efficient. In particular, NinA Transformer nearly matches the diffusion baseline (0.938 vs. 0.952 average score), while requiring an order of magnitude fewer parameters and offering up much faster inference (\autoref{fig:overall_avg}). Even the extremely compact NinA MLP (2M parameters) attains strong results (0.909 average score). These findings highlight a key advantage of NinA: high efficiency without sacrificing task success, an essential property for real-world robotic deployment where both latency and memory constraints are critical.

We next test whether NinA benefits disproportionately from FLOWER’s robotic VLM pretraining. Replacing the robotics-pretrained VLM with a generic one slightly reduces performance in both cases, but differences remain small. Interestingly, NinA MLP improves here (0.922 vs.\ 0.909), suggesting its inductive bias reduces reliance on robotics-specific VLM features. Thus, NinA generalizes effectively to different VLMs.

We also ablate the effect of removing noise injection. As shown in \autoref{tab:core}, both NinA MLP and Transformer see clear performance drops (0.880 vs.\ 0.909 and 0.896 vs.\ 0.938). This confirms noise injection is an important regularizer.

Finally, we investigate PLU augmentations \citep{kingma2018glow}. Removing PLU slightly lowers NinA Transformer (0.934 vs.\ 0.938), while NinA MLP shows mixed results. Thus, PLU provides modest but non-essential gains, whose utility may depend on NF architecture. Prior work \citep{ghugare2025normalizing} did not explore this ablation, leaving open when such augmentations are most useful. More ablations are provided in \autoref{app:abl}.
\section{Conclusion}
We presented NinA, a Normalizing Flow-based action expert for VLA models, as an efficient alternative to diffusion-based policies. NinA achieves competitive performance on the LIBERO benchmark while being substantially faster and more parameter-efficient. Beyond efficiency, the exact likelihood estimation provided by NFs opens up opportunities for integration with reinforcement learning, uncertainty modeling, and interpretability. Future work should explore scaling NinA to full VLA pretraining across diverse datasets, domains, and VLM backbones, as well as investigating its benefits for real-world robotic control.


\bibliography{b}

\begin{thebibliography}{21}
\providecommand{\natexlab}[1]{#1}
\providecommand{\url}[1]{\texttt{#1}}
\expandafter\ifx\csname urlstyle\endcsname\relax
  \providecommand{\doi}[1]{doi: #1}\else
  \providecommand{\doi}{doi: \begingroup \urlstyle{rm}\Url}\fi

\bibitem[Akimov et~al.(2022)Akimov, Kurenkov, Nikulin, Tarasov, and Kolesnikov]{akimov2022let}
Dmitriy Akimov, Vladislav Kurenkov, Alexander Nikulin, Denis Tarasov, and Sergey Kolesnikov.
\newblock Let offline rl flow: Training conservative agents in the latent space of normalizing flows.
\newblock \emph{arXiv preprint arXiv:2211.11096}, 2022.

\bibitem[Black et~al.()Black, Brown, Driess, Esmail, Equi, Finn, Fusai, Groom, Hausman, Ichter, et~al.]{black2410pi0}
Kevin Black, Noah Brown, Danny Driess, Adnan Esmail, Michael Equi, Chelsea Finn, Niccolo Fusai, Lachy Groom, Karol Hausman, Brian Ichter, et~al.
\newblock $\pi$0: A vision-language-action flow model for general robot control. corr, abs/2410.24164, 2024. doi: 10.48550.
\newblock \emph{arXiv preprint ARXIV.2410.24164}.

\bibitem[Brohan et~al.(2022)Brohan, Brown, Carbajal, Chebotar, Dabis, Finn, Gopalakrishnan, Hausman, Herzog, Hsu, et~al.]{brohan2022rt}
Anthony Brohan, Noah Brown, Justice Carbajal, Yevgen Chebotar, Joseph Dabis, Chelsea Finn, Keerthana Gopalakrishnan, Karol Hausman, Alex Herzog, Jasmine Hsu, et~al.
\newblock Rt-1: Robotics transformer for real-world control at scale.
\newblock \emph{arXiv preprint arXiv:2212.06817}, 2022.

\bibitem[Dinh et~al.(2014)Dinh, Krueger, and Bengio]{dinh2014nice}
Laurent Dinh, David Krueger, and Yoshua Bengio.
\newblock Nice: Non-linear independent components estimation.
\newblock \emph{arXiv preprint arXiv:1410.8516}, 2014.

\bibitem[Dinh et~al.(2016)Dinh, Sohl-Dickstein, and Bengio]{dinh2016density}
Laurent Dinh, Jascha Sohl-Dickstein, and Samy Bengio.
\newblock Density estimation using real nvp.
\newblock \emph{arXiv preprint arXiv:1605.08803}, 2016.

\bibitem[Ghugare \& Eysenbach(2025)Ghugare and Eysenbach]{ghugare2025normalizing}
Raj Ghugare and Benjamin Eysenbach.
\newblock Normalizing flows are capable models for rl.
\newblock \emph{arXiv preprint arXiv:2505.23527}, 2025.

\bibitem[Intelligence et~al.(2025)Intelligence, Black, Brown, Darpinian, Dhabalia, Driess, Esmail, Equi, Finn, Fusai, et~al.]{intelligence2025pi_}
Physical Intelligence, Kevin Black, Noah Brown, James Darpinian, Karan Dhabalia, Danny Driess, Adnan Esmail, Michael Equi, Chelsea Finn, Niccolo Fusai, et~al.
\newblock $\pi_{0.5}$: a vision-language-action model with open-world generalization.
\newblock \emph{arXiv preprint arXiv:2504.16054}, 2025.

\bibitem[Kim et~al.(2024)Kim, Pertsch, Karamcheti, Xiao, Balakrishna, Nair, Rafailov, Foster, Lam, Sanketi, et~al.]{kim2024openvla}
Moo~Jin Kim, Karl Pertsch, Siddharth Karamcheti, Ted Xiao, Ashwin Balakrishna, Suraj Nair, Rafael Rafailov, Ethan Foster, Grace Lam, Pannag Sanketi, et~al.
\newblock Openvla: An open-source vision-language-action model.
\newblock \emph{arXiv preprint arXiv:2406.09246}, 2024.

\bibitem[Kingma \& Dhariwal(2018)Kingma and Dhariwal]{kingma2018glow}
Durk~P Kingma and Prafulla Dhariwal.
\newblock Glow: Generative flow with invertible 1x1 convolutions.
\newblock \emph{Advances in neural information processing systems}, 31, 2018.

\bibitem[Kolesnikov et~al.(2024)Kolesnikov, Pinto, and Tschannen]{kolesnikov2024jet}
Alexander Kolesnikov, Andr{\'e}~Susano Pinto, and Michael Tschannen.
\newblock Jet: A modern transformer-based normalizing flow.
\newblock \emph{arXiv preprint arXiv:2412.15129}, 2024.

\bibitem[Liu et~al.(2023)Liu, Zhu, Gao, Feng, Liu, Zhu, and Stone]{liu2023libero}
Bo~Liu, Yifeng Zhu, Chongkai Gao, Yihao Feng, Qiang Liu, Yuke Zhu, and Peter Stone.
\newblock Libero: Benchmarking knowledge transfer for lifelong robot learning.
\newblock \emph{Advances in Neural Information Processing Systems}, 36:\penalty0 44776--44791, 2023.

\bibitem[Reuss et~al.()Reuss, Zhou, R{\"u}hle, Ya{\u{g}}murlu, Otto, and Lioutikov]{reussflower}
Moritz Reuss, Hongyi Zhou, Marcel R{\"u}hle, {\"O}mer~Erdin{\c{c}} Ya{\u{g}}murlu, Fabian Otto, and Rudolf Lioutikov.
\newblock Flower: Democratizing generalist robot policies with efficient vision-language-action flow policies.
\newblock In \emph{7th Robot Learning Workshop: Towards Robots with Human-Level Abilities}.

\bibitem[Rezende \& Mohamed(2015)Rezende and Mohamed]{rezende2015variational}
Danilo Rezende and Shakir Mohamed.
\newblock Variational inference with normalizing flows.
\newblock In \emph{International conference on machine learning}, pp.\  1530--1538. PMLR, 2015.

\bibitem[Team et~al.(2024)Team, Ghosh, Walke, Pertsch, Black, Mees, Dasari, Hejna, Kreiman, Xu, et~al.]{team2024octo}
Octo~Model Team, Dibya Ghosh, Homer Walke, Karl Pertsch, Kevin Black, Oier Mees, Sudeep Dasari, Joey Hejna, Tobias Kreiman, Charles Xu, et~al.
\newblock Octo: An open-source generalist robot policy.
\newblock \emph{arXiv preprint arXiv:2405.12213}, 2024.

\bibitem[Vaswani et~al.(2017)Vaswani, Shazeer, Parmar, Uszkoreit, Jones, Gomez, Kaiser, and Polosukhin]{vaswani2017attention}
Ashish Vaswani, Noam Shazeer, Niki Parmar, Jakob Uszkoreit, Llion Jones, Aidan~N Gomez, {\L}ukasz Kaiser, and Illia Polosukhin.
\newblock Attention is all you need.
\newblock \emph{Advances in neural information processing systems}, 30, 2017.

\bibitem[Wu et~al.(2023)Wu, Jing, Cheang, Chen, Xu, Li, Liu, Li, and Kong]{wu2023unleashing}
Hongtao Wu, Ya~Jing, Chilam Cheang, Guangzeng Chen, Jiafeng Xu, Xinghang Li, Minghuan Liu, Hang Li, and Tao Kong.
\newblock Unleashing large-scale video generative pre-training for visual robot manipulation.
\newblock \emph{arXiv preprint arXiv:2312.13139}, 2023.

\bibitem[Xiao et~al.(2024)Xiao, Wu, Xu, Dai, Hu, Lu, Zeng, Liu, and Yuan]{xiao2024florence}
Bin Xiao, Haiping Wu, Weijian Xu, Xiyang Dai, Houdong Hu, Yumao Lu, Michael Zeng, Ce~Liu, and Lu~Yuan.
\newblock Florence-2: Advancing a unified representation for a variety of vision tasks.
\newblock In \emph{Proceedings of the IEEE/CVF Conference on Computer Vision and Pattern Recognition}, pp.\  4818--4829, 2024.

\bibitem[Yang et~al.(2023)Yang, Zhang, Song, Hong, Xu, Zhao, Zhang, Cui, and Yang]{yang2023diffusion}
Ling Yang, Zhilong Zhang, Yang Song, Shenda Hong, Runsheng Xu, Yue Zhao, Wentao Zhang, Bin Cui, and Ming-Hsuan Yang.
\newblock Diffusion models: A comprehensive survey of methods and applications.
\newblock \emph{ACM computing surveys}, 56\penalty0 (4):\penalty0 1--39, 2023.

\bibitem[Zhai et~al.(2024)Zhai, Zhang, Nakkiran, Berthelot, Gu, Zheng, Chen, Bautista, Jaitly, and Susskind]{zhai2024normalizing}
Shuangfei Zhai, Ruixiang Zhang, Preetum Nakkiran, David Berthelot, Jiatao Gu, Huangjie Zheng, Tianrong Chen, Miguel~Angel Bautista, Navdeep Jaitly, and Josh Susskind.
\newblock Normalizing flows are capable generative models.
\newblock \emph{arXiv preprint arXiv:2412.06329}, 2024.

\bibitem[Zhao et~al.(2023)Zhao, Kumar, Levine, and Finn]{zhao2023learning}
Tony~Z Zhao, Vikash Kumar, Sergey Levine, and Chelsea Finn.
\newblock Learning fine-grained bimanual manipulation with low-cost hardware.
\newblock \emph{arXiv preprint arXiv:2304.13705}, 2023.

\bibitem[Zitkovich et~al.(2023)Zitkovich, Yu, Xu, Xu, Xiao, Xia, Wu, Wohlhart, Welker, Wahid, et~al.]{zitkovich2023rt}
Brianna Zitkovich, Tianhe Yu, Sichun Xu, Peng Xu, Ted Xiao, Fei Xia, Jialin Wu, Paul Wohlhart, Stefan Welker, Ayzaan Wahid, et~al.
\newblock Rt-2: Vision-language-action models transfer web knowledge to robotic control.
\newblock In \emph{Conference on Robot Learning}, pp.\  2165--2183. PMLR, 2023.

\end{thebibliography}
\bibliographystyle{iclr2025_conference}

\newpage
\appendix

\section{Additional Details}
\label{app:details}

\paragraph{Hardware.} 
All training experiments were conducted on NVIDIA H100 GPUs. 
For inference-time measurements reported in \autoref{fig:overall_avg} and \autoref{app:inference}, we additionally used a personal NVIDIA RTX 3060 Mobile GPU to provide a perspective on performance under more resource-limited settings. 

\paragraph{Codebase and Modifications.} 
Our implementation builds on the FLOWER framework \citep{reussflower}, available at \url{https://github.com/intuitive-robots/flower_vla_calvin}. 
Beyond incorporating Normalizing Flows, we introduced only two modifications: (i) the number of training epochs was increased to $100$ to account for training action experts from scratch, and (ii) the batch size was set to $80$ to accelerate training. 
For all experiments, we used Florence-2 Large \citep{xiao2024florence}, which was finetuned in the original FLOWER work, as the VLM. 

\paragraph{Hyperparameter Selection and Evaluation.} 
Hyperparameter choices were determined using the LIBERO-10 benchmark, selecting the configuration with the highest success rate at the final checkpoint after $100$ epochs. 
For evaluations on other LIBERO tasks, we consistently report the success rate of the final checkpoint as well, without additional tuning.

\begin{figure}[ht]
    \centering
    \includegraphics[width=0.95\linewidth]{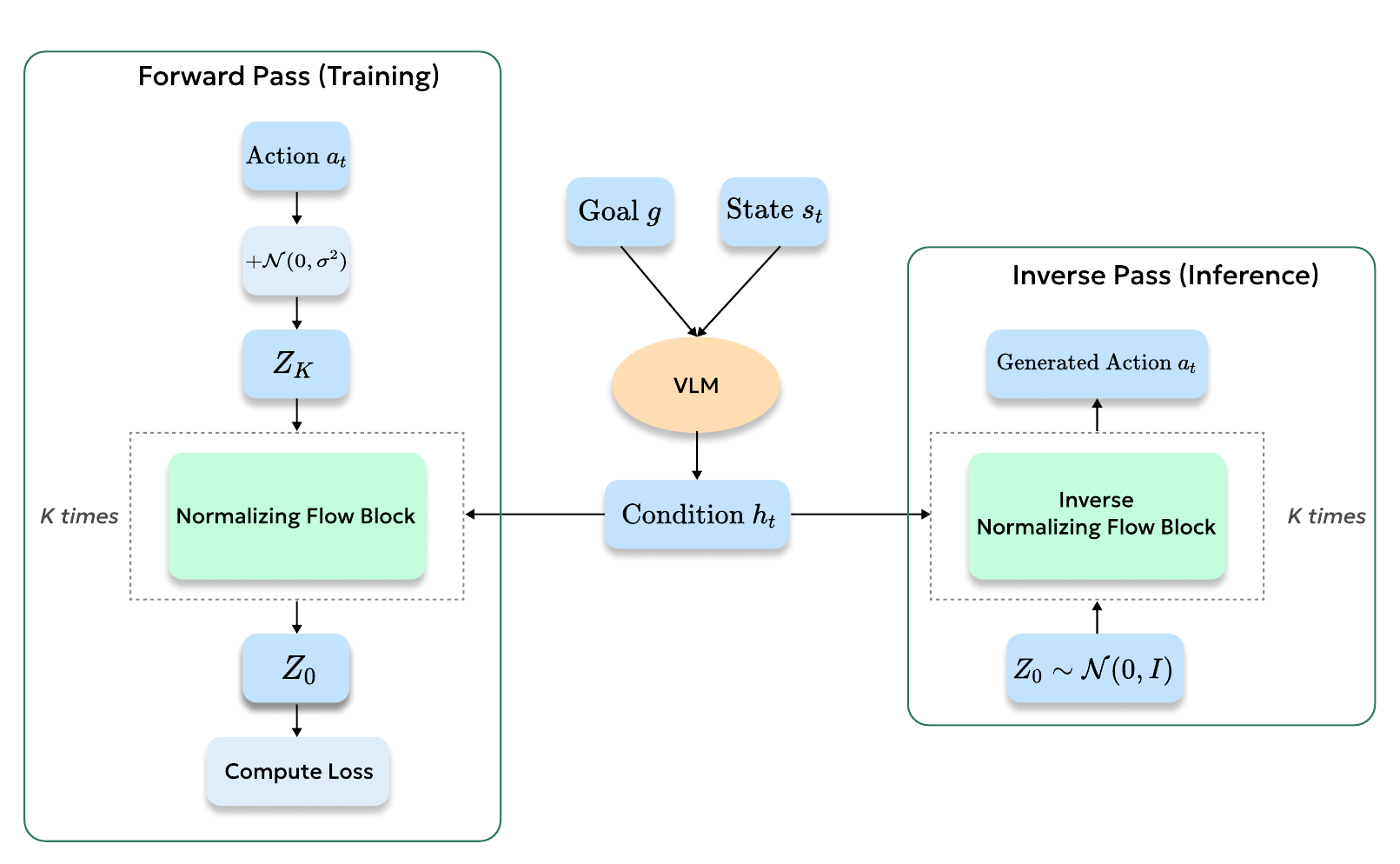}
    \caption{NinA training and inference passes, see \autoref{fig:variants} for Normalizing Flow blocks variants. See \autoref{methodology} for textual description.}
    \label{fig:passes}
\end{figure}

\begin{figure}[ht]
    \centering
    \includegraphics[width=0.7\linewidth]{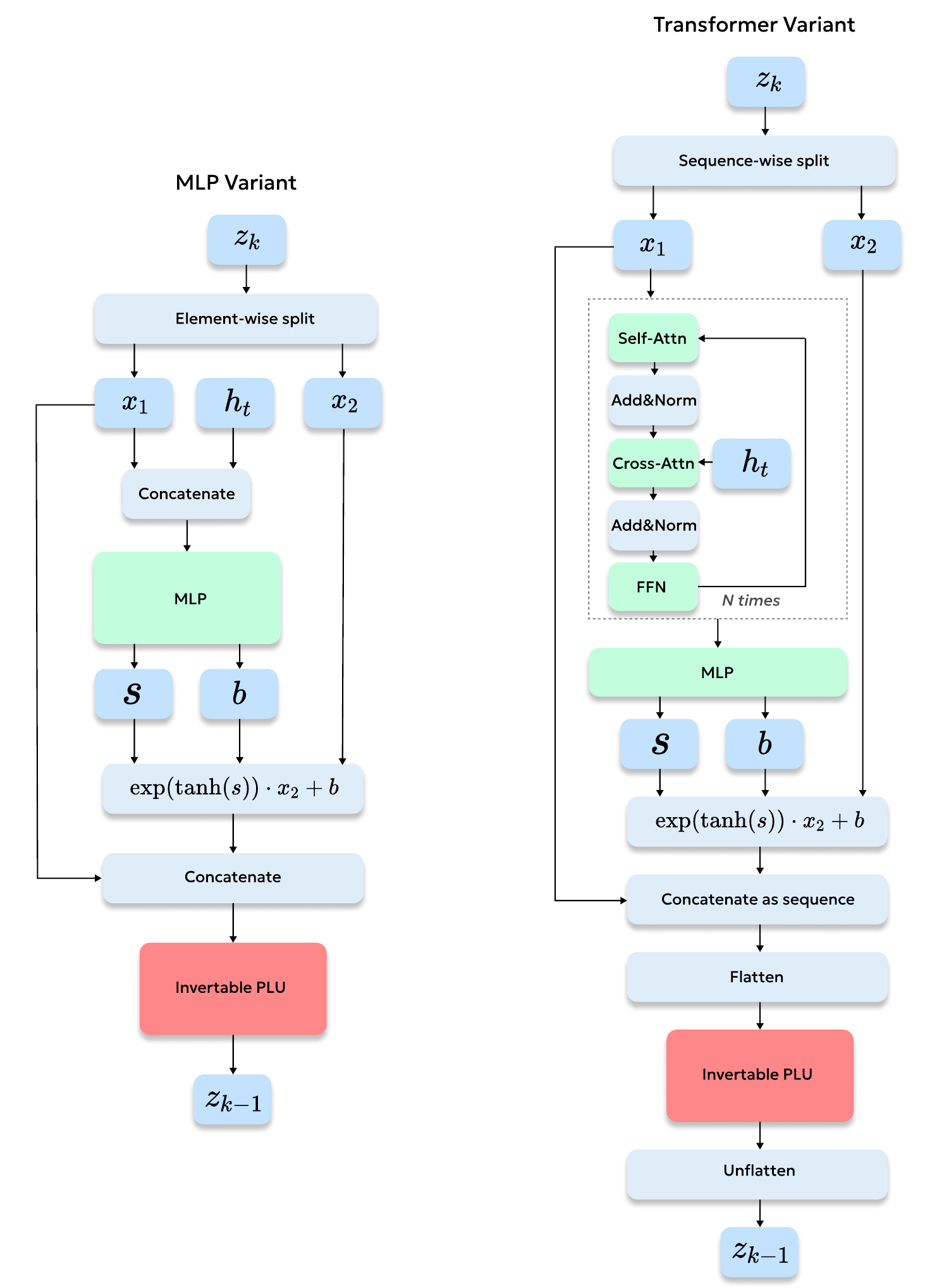}
    \caption{NinA Normalizing Flow variants schemes. See \autoref{methodology} for textual description.}
    \label{fig:variants}
\end{figure}

In \autoref{fig:passes} and \autoref{fig:variants} we present schematic illustration of our approach described in \autoref{methodology}.

\label{app:additional}

\newpage
\section{Hyperparameters}
\label{app:hps}
In this section we report the hyperparameters used for training NinA with both Transformer and MLP backbones. 
The reported values correspond to the configurations that achieved the best performance on LIBERO-10 and were therefore used for all main experiments. 
We include the flow depth ($K$), hidden dimensionality of the flow layers, noise amplitude applied to reference actions during training, and the number of layers per non-linear transformation inside each flow block $N$ (number of layers for MLP and number of stacked pairs of self- and cross-attention for Transformer).
\autoref{tab:hyperparams_transformer} summarize the settings. 

\begin{table}[h]
\centering
\caption{Hyperparameters for NinA with Transformer backbone.}
\label{tab:hyperparams_transformer}
\begin{tabular}{lcccc}
\toprule
\textbf{Model} & \textbf{Flow depth (K)} & \textbf{Flow hidden dim} & \textbf{Noise amplitude} & \textbf{N (depth per flow)} \\
\midrule
NinA MLP & 28 & 64 & 0.03 & 3 \\
NinA Transformer & 18 & 256 & 0.03 & 3 \\
\bottomrule
\end{tabular}
\end{table}

\section{Inference Time Details}
\label{app:inference}

We report the inference time per sample (in seconds) for NinA and baseline diffusion models on two hardware setups: NVIDIA H100 (server-grade) and NVIDIA RTX 3060 Mobile (consumer-grade). 
All models were implemented in PyTorch. 
Note that we only measure the action generation module, excluding the VLM forward pass, to provide a fair comparison of the generative component itself. 
Interestingly, the diffusion model with 32M parameters runs slightly slower than its 330M counterpart. 
We attribute this to changes in hidden dimensionality that likely resulted in suboptimal kernel fusion and reduced GPU utilization. 

As shown in Table~\ref{tab:inference_time}, NinA achieves \textbf{an order of magnitude faster inference} compared to diffusion models. 
This efficiency gain is consistent across both the high-end H100 GPU and the consumer-grade RTX 3060 Mobile, demonstrating that our approach is not only lightweight but also substantially more practical for real-time deployment. 

\begin{table}[h]
\centering
\caption{Inference time per sample (seconds) for different models on H100 and RTX 3060 GPUs. 
Only the action generation module is measured (VLM inference excluded). 
NinA significantly outperforms diffusion baselines in inference speed.}
\label{tab:inference_time}
\begin{tabular}{lcc}
\toprule
\textbf{Model} & \textbf{H100} & \textbf{RTX 3060} \\
\midrule
Diffusion (330M) & 0.110 & 0.163 \\
Diffusion (32M)  & 0.120 & 0.181 \\
NinA Transformer (38M) & 0.021 & 0.023 \\
NinA MLP (2M) & 0.015 & 0.019 \\
\bottomrule
\end{tabular}
\end{table}

\section{Ablations}
\label{app:abl}
We further investigate the design choices of NinA by varying the flow depth, hidden dimensionality of the flow networks, and the amount of noise added to reference actions during training. Results are reported on the LIBERO-10 benchmark. 

\begin{figure}[ht]
    \centering
    \includegraphics[width=0.95\linewidth]{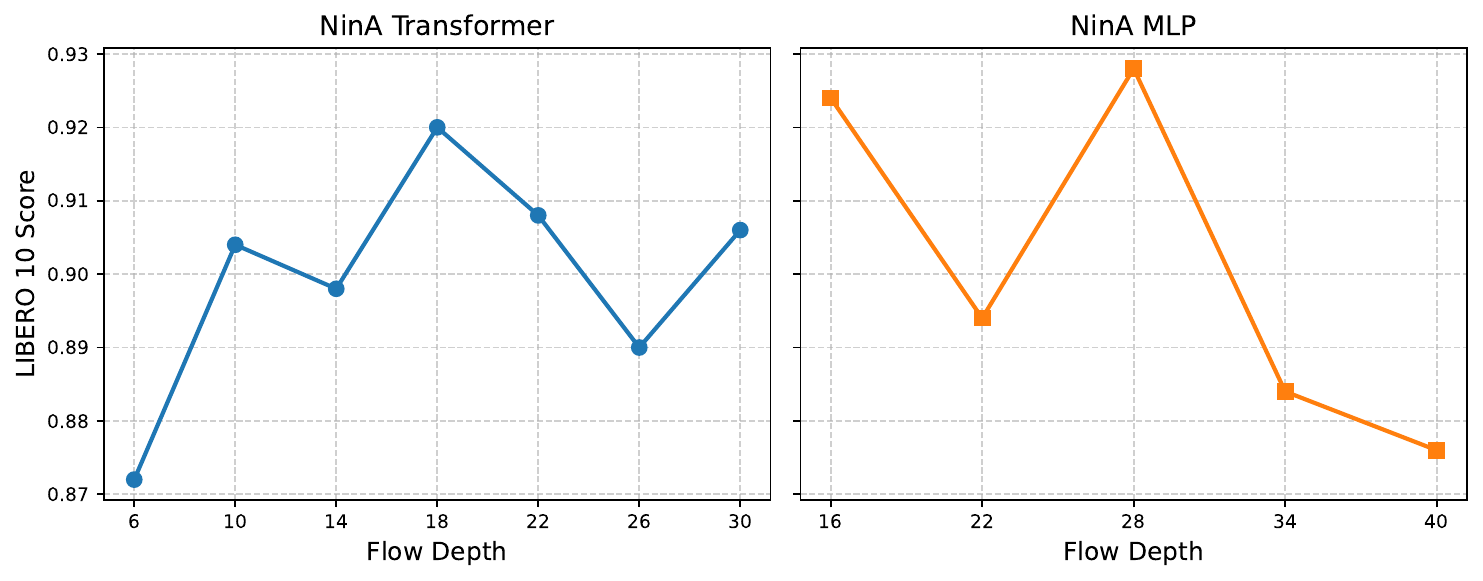}
    \caption{Ablation on the number of flows (depth) for NinA Transformer and NinA MLP on LIBERO-10.}
    \label{fig:depth_ablation}
\end{figure}

\paragraph{Flow Depth.} 
\autoref{fig:depth_ablation} compares NinA with Transformer and MLP backbones across different flow depths. 
The Transformer variant demonstrates stable performance even as the depth scales to larger values, peaking at depth $18$ (0.92) and maintaining competitive results up to depth $30$. 
In contrast, the MLP variant shows stronger fluctuations, with a peak at depth $28$ (0.928) but noticeable degradation for deeper architectures. 
These results suggest that Transformers provide a more scalable backbone for NinA, motivating their use when considering larger models.

\begin{figure}[ht]
    \centering
    \includegraphics[width=0.95\linewidth]{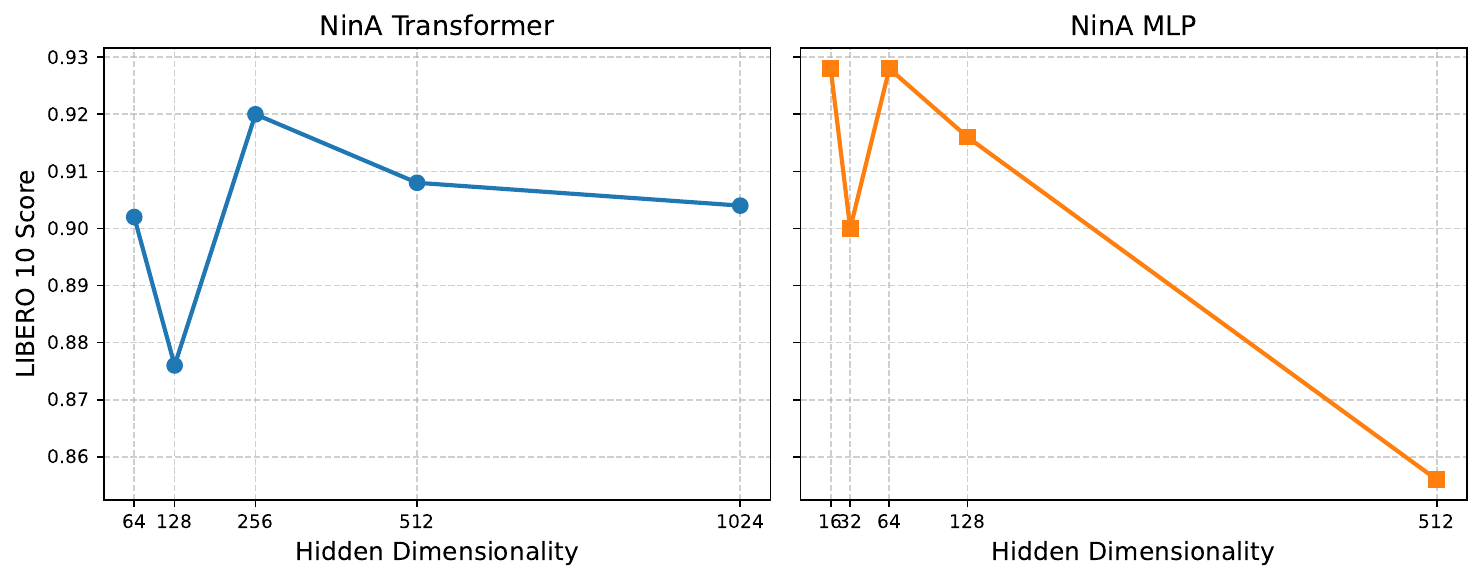}
    \caption{Ablation on the hidden dimensionality of flows for NinA Transformer and NinA MLP on LIBERO-10.}
    \label{fig:hidden_ablation}
\end{figure}

\paragraph{Hidden Dimensionality.} 
As shown in \autoref{fig:hidden_ablation}, Transformer-based NinA achieves its best performance with a hidden dimension of $256$ (0.920), while the MLP variant performs well for smaller hidden sizes, peaking at $16$ and $64$ (0.928). 
This indicates that Transformers benefit from moderately larger hidden sizes, whereas MLPs perform strongly even with compact representations.

\begin{figure}[ht]
    \centering
    \includegraphics[width=0.95\linewidth]{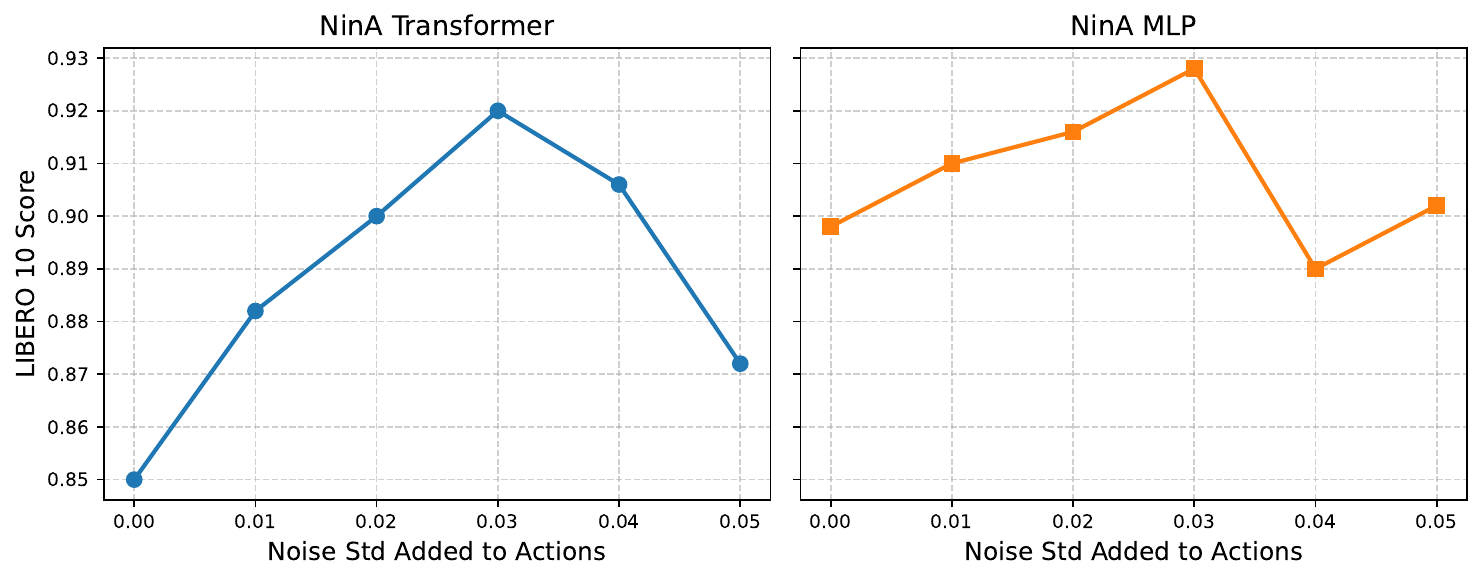}
    \caption{Ablation on Gaussian noise injected into reference actions for NinA Transformer and NinA MLP on LIBERO-10.}
    \label{fig:noise_ablation}
\end{figure}

\paragraph{Noise Injection.} 
In \autoref{fig:noise_ablation}, we study the effect of adding Gaussian noise to the reference actions. 
For both Transformer and MLP backbones, performance improves with small amounts of noise and peaks at $\sigma=0.03$ 
(0.920 for Transformer, 0.928 for MLP), after which it decreases as noise grows larger. 
Interestingly, the MLP variant shows slightly higher robustness across the noise range, maintaining strong performance even at $\sigma=0.05$. 
These results suggest that moderate stochasticity during training serves as a useful regularizer and that MLPs may benefit more consistently from this effect. 

\end{document}